\documentclass[10pt,twocolumn,letterpaper]{article}

\usepackage{iccv}
\usepackage{times}
\usepackage{epsfig}
\usepackage{graphicx}
\usepackage{amsmath}
\usepackage{amssymb}
\usepackage{subfigure}

\usepackage[breaklinks=true,bookmarks=false]{hyperref}

\iccvfinalcopy %

\def\iccvPaperID{****} %

\begin{document}

\title{Learning to Find Correlated Features by Maximizing Information Flow in Convolutional Neural Networks}

\author{Wei Shen  \qquad Fei Li \qquad Rujie Liu \\
Fujitsu Research \& Development Center, Beijing, China.\\
{\tt\small \{shenwei, lifei, rjliu\}@cn.fujitsu.com}
}

\maketitle

\begin{abstract}
Training convolutional neural networks for image classification tasks usually causes information loss. Although most of the time the  information lost is redundant with respect to the target task, there are still cases where discriminative information is also discarded. 
For example, if the samples that belong to the same category have multiple correlated features, the model may only learn a subset of the features and ignore the rest. This may not be a problem unless the classification in the test set highly depends on the ignored features. We argue that the discard of the correlated discriminative information is partially caused by the fact that the minimization of the classification loss doesn't ensure to learn the overall discriminative information but only the most discriminative information.
To address this problem, we propose an information flow maximization (IFM) loss as a regularization term to find the discriminative correlated features. With less information loss the classifier can make predictions based on more informative features. We validate our method on the shiftedMNIST dataset and show the effectiveness of IFM loss in learning representative and discriminative features.
\end{abstract}

\section{Introduction}
Usually a classification model is  trained with a softmax loss which is quite successful in many scenarios. This loss typical helps the model learn discriminative features for the target task and ignore the irrelevant features. However, if there are several discriminative features that are correlated within a category, the model may choose the most discriminative feature (\eg color and texture) and ignore the rest (\eg object structure). 
The reason is that the most discriminative features are those that make the steepest descend in loss function. As the training continues, those features will dominate the final feature representation and the rest discriminative features will be discarded as well as the irrelevant features. Similar evidence can also be found in recent study on ImageNet-trained CNNs. It shows that those models are biased towards texture rather than object shape~\cite{geirhos2018imagenet}.
Learning partial discriminative features does not make the most of the dataset and thus reduces the generalization capability of the model.

Information theory has been widely used to improve the representation capability of deep neural networks~\cite{chen2016infogan,gabrie2018entropy,hjelm2018learning,shwartz2017opening,tishby2000information,zhao2017infovae}. 
In this work, we focus on how to apply mutual information to find correlated features in image classification tasks.  
According to Data Processing Inequalities (DPI), the mutual information between the input data and the hidden layers are decreasing as the layer goes deeper~\cite{tishby2015deep}. The main idea of information bottleneck (IB) trade off is that we can try to minimize the mutual information between the input data and the hidden representation and maximize the mutual information between the hidden representation and the label to find the optimal achievable representations of the input data~\cite{tishby2015deep}. In contrast, we find that when discriminative features are correlated, the \textit{maximization} instead of minimization of mutual information between hidden representations can provide extra benefits for representation learning. We call this strategy information flow maximization (IFM) which is achieved by estimating and maximizing the mutual information between convolutional layers simultaneously. IFM is implemented using a multi-layer fully connected neural network and it serves as a plugin in the conventional CNNs in the training stage. In the test stage, the IFM block is removed and thus there is no extra computation cost.

\section{Related work}
There are many work concentrating on information maximization for deep networks. 
In ~\cite{chen2016infogan}, Chen \etal introduce InfoGAN which is a generative adversarial network that maximizes the mutual information between a small subset of the latent variables and the observation. In~\cite{belghazi2018mine}, Belghazi \etal present a Mutual Information
Neural Estimator (MINE) that estimates mutual information between high dimensional continuous random variables by gradient descent over neural networks.
In~\cite{hjelm2018learning}, Hjelm \etal introduce Deep InfoMax (DIM) to maximize mutual information between a representation and the output of a deep neural network encoder to improve the representation's suitability for downstream tasks. 
In~\cite{jacobsen2018excessive}, Jacobsen \etal propose an invertible network architecture and an alternative objective that extract overall discriminative knowledge in the prediction model.

The difference between our work and ~\cite{hjelm2018learning} is that we are concentrating on maximizing the mutual information between adjacent layers so that the information loss can be reduced while \cite{hjelm2018learning} tries to maximize the mutual information between the final representation and the output convolutional feature maps.
The work in~\cite{jacobsen2018excessive} is closely related to our work, the main difference is that we apply IFM blocks instead of flow-based models to reduce the information loss. 

\section{Method}
\begin{figure}[h]
\begin{center}
\includegraphics[width=0.99\linewidth]{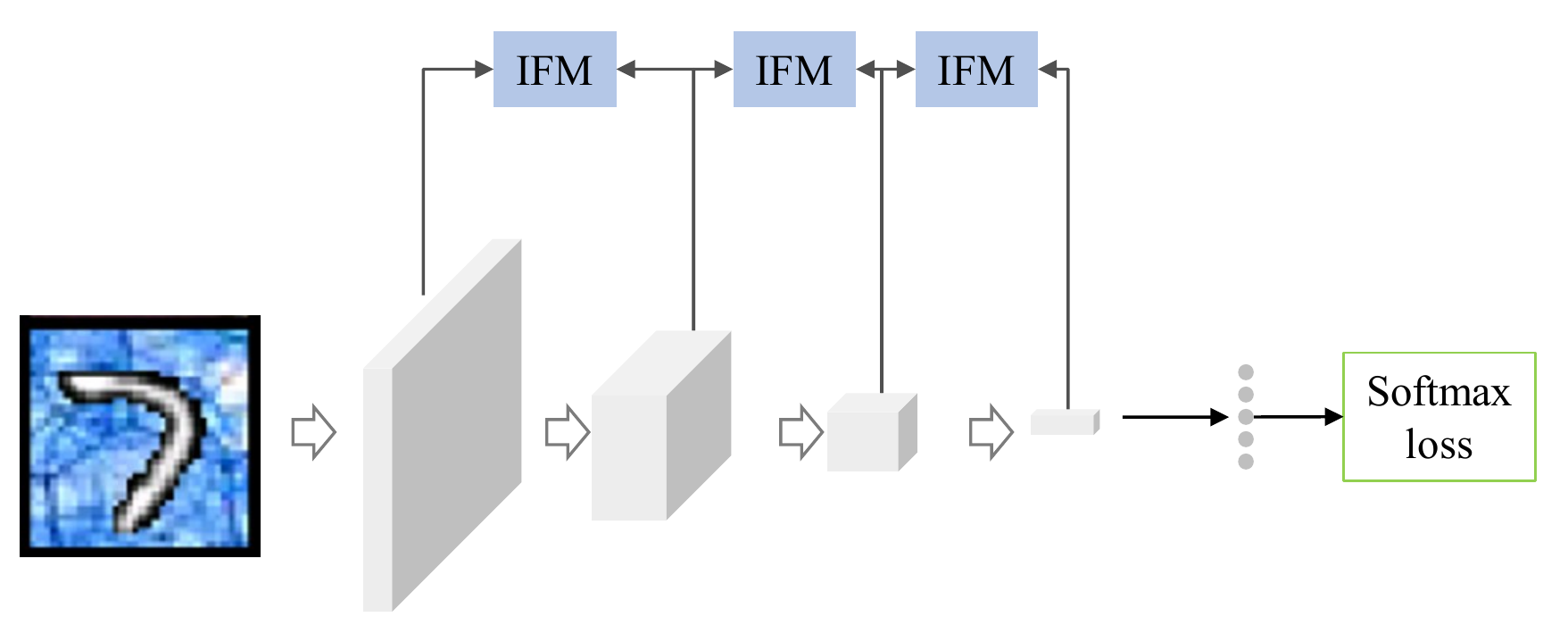} 
\end{center}
\caption{The training pipeline of the proposed method. In the training stage, the model is trained to minimize the classification loss and maximize the information flow between layers.}
\label{fig:pipeline}
\end{figure}

The pipeline of the proposed method is shown in Figure~\ref{fig:pipeline}. The backbone network is a vanilla convolutional neural network. The IFM blocks are plugged in between adjacent convolution layers. Note that the IFM blocks are only used in the training stage. In the test stage, those IFM blocks is removed so that there are no extra computation cost. 

\subsection{Mutual information estimation}
In oder to be self-contained, in this section we will introduce how to estimate mutual information.
Formally, the mutual information is calculated as
\begin{equation}
\begin{split}
I(X,Z)&=\sum\limits_{z\in Z}\sum\limits_{x\in X}p(x,z)\text{log}\frac{p(x,z)}{p(x)p(z)}\\
&=KL(p(x,z)||p(x)p(z)),
\end{split}
\label{eq:MI_KL}
\end{equation}
where $X$ and $Z$ are two random variables. $p(x,z)$ is the joint probability mass function of $X$ and $Y$. $p(x)$ and $p(z)$ are the marginal probability mass functions of $X$ and $Z$ respectively.
From Equation~\ref{eq:MI_KL}, we can find that maximizing the mutual information between $X$ and $Z$ is equivalent to maximizing the Kullback-Leibler divergence between the joint distribution of $p(x,y)$ and the product of marginal distribution of $p(x)$ and $p(z)$. 

Following ~\cite{nowozin2016f},the general form of f-divergence can be approximated by
\begin{equation}
D(P||Q) = \sup\limits_{T \in \mathcal T}(\mathbb{E}_P[T(X)] - \text{log}\mathbb{E}_Q[f^*(T(X))]),
\end{equation}
where $P$ is the joint distribution $p(x,z)$ and $Q$ is the product of marginal distribution $p(x)p(z)$. $\mathcal T$ is an arbitrary class of functions $T: X\mapsto \mathbb R$ and $f^*$ is the convex conjugate function of the generator function $f$. 
Since $D(P||Q)$ can be approximated by the supremum of the difference between two expectations, we can choose to maximize 
\begin{equation}
F(\omega) = \mathbb{E}_P[T_\omega(X)] - \text{log}\mathbb{E}_Q[f^*(T_\omega(X))]),
\end{equation}
where $T_\omega$ is a neural network parametrized by $\omega$. More specifically, $T_\omega$ can be represented in the form $T_\omega (x) = g_f(V_\omega (x))$ where $g_f$ is specific to the f-divergence used. 
Since Kullback-Leibler divergence is not upper bounded, we use Jensen-Shannon divergence as a surrogate to estimate the mutual information. 
Thus, we can replace $f^*(t)$ with $-log(2-e^t))$ and choose  $g_f(v)=log(2)-log(1 + e^{-v})$. Then we obtain
\begin{equation}
F(\omega) = \mathbb{E}_P[\frac{1}{1+e^{-V_\omega (x)}}] - \text{log}\mathbb{E}_Q[1-\frac{1}{1+e^{-V_\omega (x)}}]).
\end{equation}
Let $\sigma (v) = \frac{1}{1+e^{-v}}$. We have
\begin{equation}
F(\omega) = \mathbb{E}_P[\sigma(V_\omega (x))] - \text{log}\mathbb{E}_Q[1-\sigma(V_\omega (x))]).
\label{eq:last}
\end{equation}
$\sigma(V_\omega (x))$ is represented by network D in Figure~\ref{fig:D}. From Equation~\ref{eq:last}, we can find that the maximization of $F(\omega)$ will result in the network D outputting one for samples from the joint distribution and zero for samples from the product of the marginal distributions.

\subsection{Constructing sample pairs}
\begin{figure}[h]
\begin{center}
\subfigure[]{
  \label{fig:upsample}
  \includegraphics[width=0.45\linewidth]{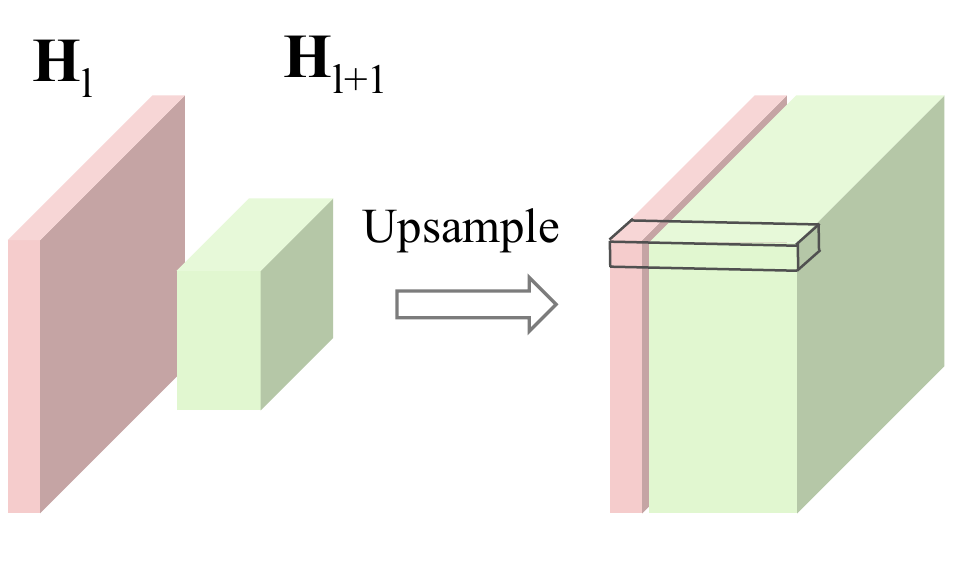}}
\hfill
\subfigure[]{
  \label{fig:D}
  \includegraphics[width=0.45\linewidth]{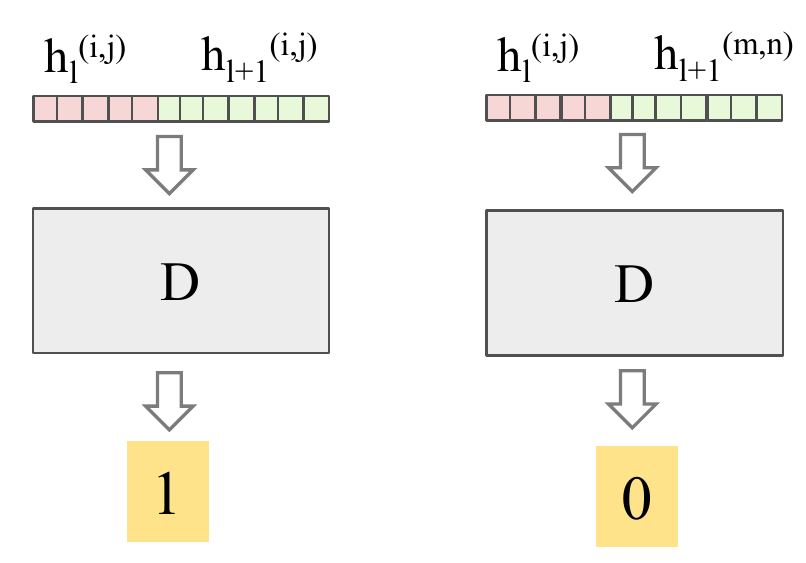}}
\end{center}
\caption{Constructing sample pairs from the joint distribution and the product of marginal distributions. $H_{l}$ and $H_{l+1}$ are two adjacent convolutional feature maps. We upsample $H_{l+1}$ to the size of  $H_{l}$ in (a). In (b), we feed sample pairs from both distributions to network D. The superscript indicates the spatial location on feature maps.}
\label{fig:MIE}
\end{figure}

In Equation~\ref{eq:last}, we still need to estimate two expectations. In the first term the samples are sampled from the joint distribution and in the second term the samples are sampled from the product of marginal distributions. Since we are estimating the mutual information between adjacent convolutional layers $H_l$ and $H_{l+1}$, we firstly resize $H_{l+1}$ to the size of $H_l$. Then sampling from the joint distribution $p(h_l, h_{l+1})$ could be achieved by sampling feature vectors at the same spatial location on the convolutional feature maps. For sampling from the second distribution, we can firstly sample a random feature vector from $H_l$ and then randomly sample another feature vector from $H_{l+1}$. For each sample pair, the two feature vectors are concatenated as a single vector. The details are shown in figure~\ref{fig:MIE}. $\sigma(V_\omega (x))$ is represented by network D and the maximization of Equation~\ref{eq:last} will optimize network D to distinguish the sample pairs from the two distributions.

\subsection{Information flow maximization}
When stacking convolutional layers, we are potentially losing information. According to DPI, we have $I(X,H_0) < I(X,H_1) < ... < I(X,H_n)$. Suppose we are given a training dataset for classification and the data representation can be decomposed into three disentangled features $z_{id0}, z_{id1}$ and $z_{v}$ where $z_{id0}$ and $z_{id1}$ can be used for classification and $z_{v}$ describes some random variations that are shared across categories. Ideally, we can used these three features to perfectly reconstruct the input data. When we are training a model for the target classification task, the information about $z_{v}$ will be gradually discarded from the information flow which is as expected. However, if the classification task is biased towards one of the id features, say $z_{id0}$, we may unexpectedly lose the information of  $z_{id1}$ as well in the information flow.
This is because during model training $\frac{\partial L}{\partial z_{id0}}$ will be much larger than $\frac{\partial L}{\partial z_{id1}}$. $z_{id0}$ will get more and more strengthened than $z_{id1}$. Finally, our model will rely only on $z_{id0}$ for classification. This behavior undermines the generalization capability of our model especially when the test task depends on $z_{id1}$ for classification. 

In order to reduce the loss of information in the information flow, we propose to maximize the mutual information between adjacent convolutional layers. The entire objective function is 
\begin{equation}
L = L_{clf}-\sum_l^LF_l(\omega),
\end{equation}
where $L_{clf}$ is the classification loss (\eg the softmax loss) and $L$ is the number of layers that used to calculate the information flow.

Although some task-irrelevant information may also be involved in the final representation, the training process will let the discriminative information dominate the representation. Thus the classifier can make predictions based on more informative features.

\section{Experiments}
\subsection{Dataset}
\begin{figure}[h]
\begin{center}
\subfigure[]{
  \label{fig:sM_train}
  \includegraphics[width=1.0\linewidth]{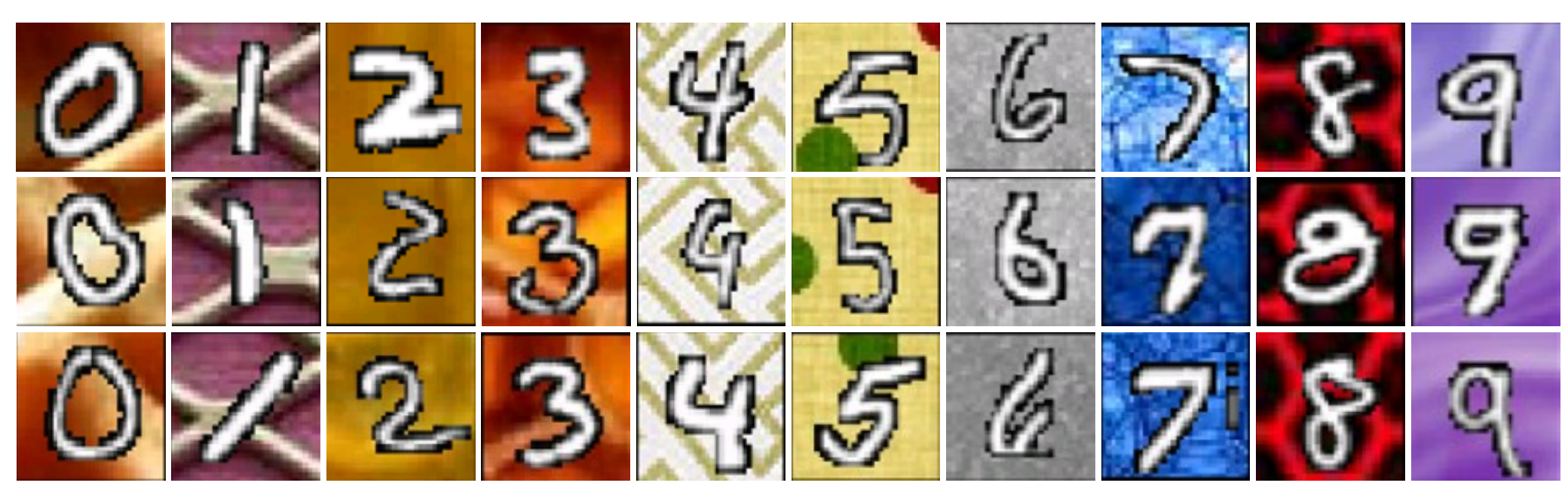}}
\hfill
\subfigure[]{
  \label{fig:sM_test}
  \includegraphics[width=1.0\linewidth]{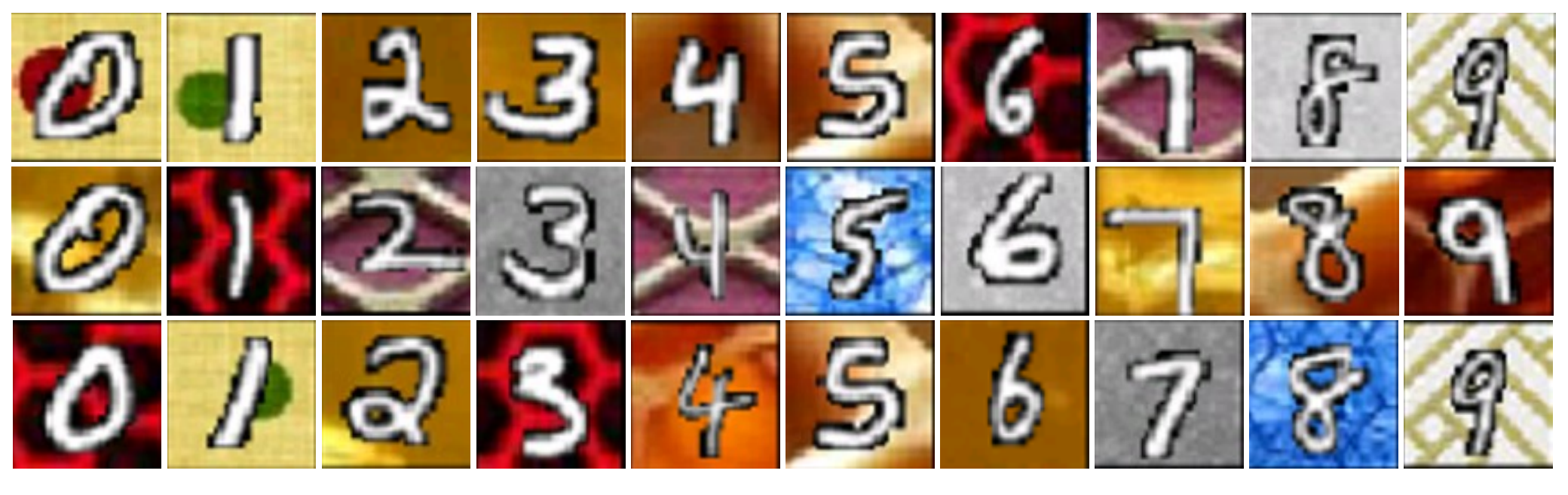}}
\end{center}
\caption{Some training examples (a) and test samples (b) from the shiftedMNIST dataset.}
\label{fig:shiftedMNIST}
\end{figure}

The dataset we used in evaluation is the shiftMNIST dataset introduced in ~\cite{jacobsen2018excessive}. It is a modified version of MNIST dataset. For the ten digits, ten texture images are randomly selected from a texture dataset~\cite{cimpoi2014describing} and applied on the digit as its background. We split one-fifth of the original MNIST training set to construct the validation set.
In the training set, each digit is associated with a fixed type of texture. For example,  for digit 1, its background patch is sampled from texture 1, and for digit 2, its background patch is sampled from texture 2, etc. However, in the validation set and test set, the digit is associated with a random texture.  In other words, the texture id and the digit id are the same for given a training image while they are not necessarily the same for a given validation or test image.  Some examples from the shiftedMNIST dataset are shown in Figure~\ref{fig:shiftedMNIST}. 

\subsection{Implementation details}
\begin{table}
\begin{center}
\begin{tabular}{c|c}
\hline
Layer & Network details\\
\hline\hline
conv1 & Conv(32,3,3)-BN-leakyReLU \\
- & Maxpool(2,2) \\
conv2 & Conv(64,3,3)-BN-leakyReLU \\
- & Maxpool(2,2) \\
conv3 & Conv(128,3,3)-BN-leakyReLU \\
- & Maxpool(2,2) \\
conv4 & Conv(128,3,3)-BN-leakyReLU \\
- & Maxpool(2,2) \\
$f$ & FC(128$\times$2$\times$2, 10) \\
\hline
\end{tabular}
\end{center}
\caption{Details of the classification network. \lq\lq Conv($c$,3,3)" indicates that there are $c$ convolution kernels with size $3\times 3$. \lq\lq Maxpool(2,2) " means max-pooling with stride 2 and the pooling window size is $2\times 2$. \lq\lq BN " indicates batch normalization~\cite{Ioffe2015}. All leakyReLUs share the same ratio of 0.2 in the negative region. \lq\lq FC " is the fully connected layer.}
\label{tb:net_clf}
\end{table}

\begin{table}
\begin{center}
\begin{tabular}{c|c}
\hline
Layer & Network details\\
\hline\hline
fc1 & FC( $N$, 256)-BN-leakyReLU \\
fc2 & FC(256, 128)-BN-leakyReLU \\
fc3 & FC(128, 64)-BN-leakyReLU \\
fc4 & FC(64, 1)-sigmoid \\
\hline
\end{tabular}
\end{center}
\caption{Details of the network D. $N$ indicates the input dimension of the concatenated feature vector.}
\label{tb:net_dis}
\end{table}

The details of the classification network and the network D are shown in Table~\ref{tb:net_clf} and Table~\ref{tb:net_dis}. The learning rate is 0.01. Mutual information is estimated for (conv1, conv2), (conv2, conv3) and (conv3, conv4). For each pair of the convolutional feature maps, the upsampling step uses nearest neighbor interpolation. The size of the input image is 32$\times$32. Both the classification network and the network D are trained in an end-to-end way simultaneously. 

\subsection{Evaluation protocol}
For the shiftedMNIST dataset, one may argue that only the digit feature should be considered as the correct feature for label prediction in the training set. However, as stated in ~\cite{ilyas2019adversarial}, the digit features can be viewed as a kind of human prior. For our model, it does not have such a prior so that both the digit feature and the texture feature may be viewed as the discriminative features. It leaves to the optimization dynamics to choose which feature as the final predictor. Note that the digit label and texture label are identical for a given training image. In the training stage, we select models with best digit validation accuracy and best texture validation accuracy to observe how the optimization dynamics influences the knowledge learning. The optimal test classification accuracy should be around 50\% since the classification model is not aware of whether the test task is a digit classification task or a texture classification. So it should learn both features equivalently.

\subsection{Results}
The classification results are shown in Table~\ref{tb:res}. In this section, we train a baseline model without IFM blocks. Model $Baseline_{Digit}$ achieves best digit validation accuracy and the model $Baseline_{Texture}$ achieves best texture validation accuracy. 
The test accuracies are shown in the first two rows in Table~\ref{tb:res}. For both models, the prediction accuracies on the digit are slightly above 10\% which is quite similar to random guess. It means that both models ignore the digit structure as the discriminative feature. The prediction accuracies on texture are above 95\% which means the final representations are dominated by the texture features. The results of the baseline models demonstrate that if the model is trained in the vanilla way it only learns partial discriminative features and ignore other correlated features. In this experiments, the baseline models are sensitive to texture features which is in accordance with the observations in ~\cite{geirhos2018imagenet}.

The benefit of applying IFM is shown in the bottom two rows. It can be found that the classification accuracies of digit are much higher than that of the baseline model. It indicates that our models indeed learn the digit structure as the discriminative feature. 
The reason for the test digit accuracy of $ours_{Texture}$ being lower than that of $ours_{Digit}$ is that digit structure features are more difficult to learn than texture features. When the texture features are well learned (with high texture validation accuracy), the learning of digit features may still be halfway.
$ours_{Digit}$ also outperforms the model in ~\cite{jacobsen2018excessive} which is a flow based model with no information loss. It implies that our IFM can be potentially viewed as an alternative way to flow-based models to reduce information loss in deep networks.

\begin{table}[]
\begin{center}
\begin{tabular}{c|cc}
\hline
Model & acc (digit)  & acc (texture)  \\ \hline
$Baseline_{Digit}$ & 12.44\%  & 95.07\% \\
$Baseline_{Texture}$ & 12.05\%  & 96.44\% \\
iCE fi-RevNet~\cite{jacobsen2018excessive} & 40.01\%  & - \\
\hline
$ours_{Digit}$  & 54.54 \%         & 40.41\%         \\ 
$ours_{Texture}$  & 31.78 \%         & 69.00\%         \\ 
\hline
\end{tabular}
\end{center}
\caption{Test classification accuracy from the baseline model, iCE fi-RevNet and ours. }
\label{tb:res}
\end{table}

\section{Conclusion}
In this work, we propose to maximize the information flow in convolutional neural networks as a kind of regularization term. The benefit of this regularization is that we can find correlated features that are difficult to be disentangled. Thus, the learned representations are more informative and generalizable than representations learned in conventional training without this information regularization term. 
Our future work will focus on how to apply the proposed information flow maximization on natural image classification tasks. 

{\small
\bibliographystyle{ieee}
\bibliography{egbib}
}

\end{document}